\documentclass{article}
\usepackage[margin=2.2cm]{geometry}
\usepackage[utf8]{inputenc}
\usepackage{graphicx}
\usepackage{color}
\usepackage{subcaption}
\usepackage[hidelinks]{hyperref}
\usepackage{url}
\usepackage{multirow}
\usepackage[font=sf]{caption}
\usepackage{amsmath} 
\usepackage{amssymb}  
\usepackage[font={footnotesize}]{caption}
 \usepackage{siunitx}
 \usepackage{booktabs}
 \usepackage{paralist}
 \usepackage[shortlabels]{enumitem}

\graphicspath{ {./images/} }
\usepackage[
backend=biber,
style=numeric,
sorting=none
]{biblatex}
\newcommand*{\figref}[2][]{%
  \hyperref[{fig:#2}]{%
    Figure~\ref*{fig:#2}%
    \ifx\\#1\\%
    \else
      \,#1%
    \fi
  }%
}

\newcommand*{\tableref}[2][]{%
  \hyperref[{table:#2}]{%
    Table~\ref*{table:#2}%
    \ifx\\#1\\%
    \else
      \,#1%
    \fi
  }%
}

\newcommand*{\equref}[2][]{%
  \hyperref[{eq:#2}]{%
    Equation~\ref*{eq:#2}%
    \ifx\\#1\\%
    \else
      \,#1%
    \fi
  }%
}

\begin{document}
\section*{\textsf{The Bridge between Xsens Motion-Capture and Robot Operating System (ROS): Enabling Robots with Online 3D Human Motion Tracking}}

\subsubsection*{\textsf{Mattia Leonori$^1$, Marta Lorenzini$^1$, Luca Fortini$^{*1}$, Juan M. Gandarias$^1$, Arash Ajoudani$^1$}}
$^1$Human-Robot Interfaces and Interaction, Istituto Italiano di Tecnologia, Genoa, Italy.\\
$^*$Corresponding author: \href{mailto:mattia.leonori@iit.it}{mattia.leonori@iit.it}


\section*{\textsf{Introduction}}
\label{Sec::introduction}
With the growing numbers of intelligent autonomous systems in human environments, the ability of such systems to perceive, understand, and anticipate human behavior becomes increasingly important  \cite{rudenko2020human}. In particular, the constant and accurate tracking of human motion is a key skill for robots to coexist, interact, cooperate with, or imitate humans \cite{field2009motion}. The resulting applications are countless, ranging from surveillance, navigation, and teleoperation, to those involving physical human-robot interaction and collaboration.

Essentially, motion can be recorded by tracking the precise position and orientation of points of interest at high frequency. Different physical principles can be exploited for this purpose. Resulting technologies vary from the multiplexed reading of orthogonal magnetic fields from inductive coils, accelerometers~\cite{lee2001real}, and gyroscopes~\cite{sakaguchi1996human}, intensity of ultrasonic pulses~\cite{marquet2011optimal}, or reconstruction of the position of visible markers detected with multiple cameras~\cite{kolahi2007design} as well as marker-less artificial intelligence (AI)-driven solutions~\cite{fortini2023markerless}. This results in systems with diverse accuracy and capabilities but also weaknesses, such as occluded trackers, constrained motion, or magnetic disturbances. As a result, for each one of the above-mentioned applications, a different sensor technology may be more suitable.

When portability is a key requirement, the latest releases of inertial motion-capture systems may offer an optimal solution. They do not require any stationary units (e.g. transmitter, receiver, or cameras) and thus can be used in extended spaces both indoors and outdoors. They are wearable, lightweight, and pose no hindrance to human movement, allowing the users to carry out their tasks without any constraints or discomfort. Furthermore, they are easily transportable and they can be used for prolonged periods of time (long battery life). Above all, Xsens MVN \cite{schepers2018xsens}, the second-generation wireless inertial-magnetic motion-capture system by Movella\texttrademark, enables real-time 3D kinematic applications with multiple reduced-sized motion trackers by providing highly accurate orientation through an unobtrusive setup. 

Let's suppose we want to track human motion with the Xsens MVN system to generate inputs for robot controllers and planners from human intentions, trajectories, or actions for a specific application. 
Xsens software MVN Analyze/Animate enables the online streaming of 3D kinematic data in both UDP and TCP formats. On the other hand, the most employed middleware suite for building robotics applications is the so-called Robot Operating System (ROS), an open-source set of libraries and tools for robot software development.

What if Xsens and ROS could easily communicate with each other? This would extremely facilitate the implementation of algorithms to feed robot controllers and planners with online information directly extracted from human motion and allow humans to share data with the robot by using the very language of robots.
This document introduces the bridge between the leading inertial motion-capture systems for 3D human tracking and the most used robotics software framework. 3D kinematic data provided by Xsens are translated into ROS messages to make them usable by robots and a Unified Robotics Description Format (URDF) model of the human kinematics is generated, which can be run and displayed in ROS 3D visualizer, RViz.

\section*{\textsf{Framework}}
\label{Sec::methods}
\begin{figure}[ht]
    \centering
    \includegraphics[width=\textwidth,trim={0 4cm 0 4cm},clip]{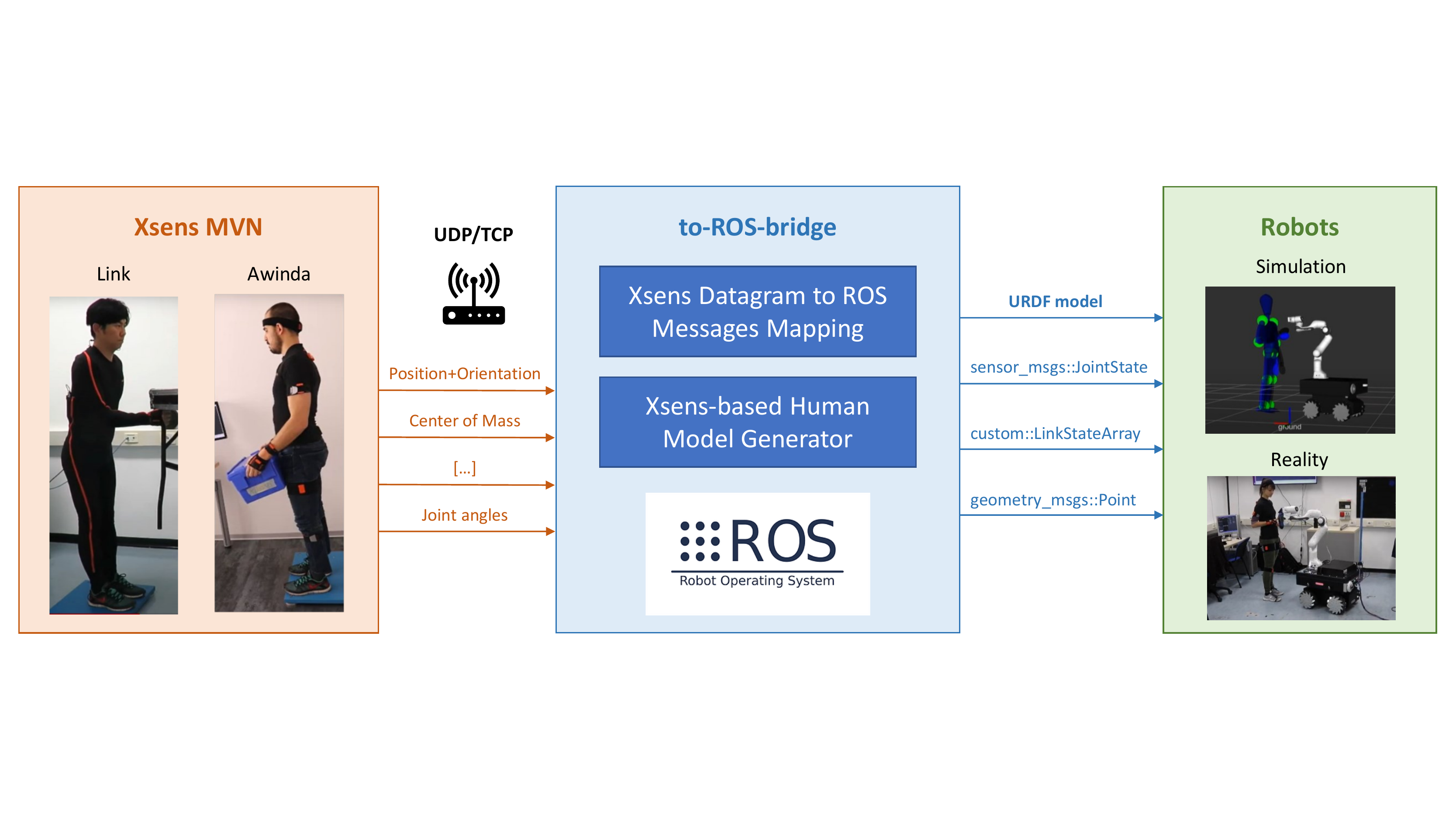}
    \caption{Overview of the framework to map Xsens 3D kinematic data to ROS-compatible paradigms.}
    \label{fig:framework}
\end{figure}

In this section, the main components of the Xsens-to-ROS bridge, depicted in Figure \ref{fig:framework}, are illustrated. First, an overview of the Xsens MVN system and the Robot Operating System (ROS), which are the foundation of the proposed framework, is provided. Then, the functioning of the bridge is described.

\subsection*{Xsens MVN}
The Xsens MVN is an easy-to-use, cost-efficient inertial system for full-body human motion capture. MVN is based on Xsens state-of-the-art miniature inertial sensors and wireless communication solutions combined with advanced sensor fusion algorithms, using assumptions of biomechanical models. MVN is a completely portable system, there are no limitations in measurement volume (apart from the wireless range). Examples of fields of use are biomechanics~\cite{schepers2007ambulatory}, sport~\cite{kruger2009biomechanical}, rehabilitation~\cite{zhou2008human}, ergonomics~\cite{maurice2019human}, and human-robot interaction~\cite{van2020predicting, rapetti2023control} as well as 3D Animation and virtual reality, training and simulation~\cite{held2020augmented}.   

In particular, the MVN suit is provided with 17 miniature inertial measurement units (IMUs) called MTx/MTw. IMUs integrate rate gyroscopes to measure 3D angular velocities, accelerometers to measure 3D accelerations (including gravity), magnetometers to measure 3D earth magnetic field, as well as atmospheric pressure using a barometer. Both a lycra suit, i.e. MVN Link, or stretchable straps, i.e. MVN Awinda, (see Figure \ref{fig:framework}, left block) are available as mounting systems to support the sensors (MTx and MTw, respectively). While MTx requires a cable system, MTw are wireless trackers. Some specifications about MVN Link/Awinda and the trackers are listed in Table \ref{table:XsensIMUspec}. 
\begin{table}[ht]
	\centering
	\caption{Xsens MVN specifications for Link, Awinda and trackers \cite{xsensmanual}}
	\label{table:XsensIMUspec}
	\resizebox{\textwidth}{!}{\begin{tabular}{ccc}
		\toprule
		\textbf{Specification} & \textbf{MVN Link} & \textbf{MVN  Awinda}\\ \midrule
		\textit{Update rate}\footnotemark & $240$ Hz & $60$ Hz \\
		\textit{Range} & $150$ m & $50$ m \\
		\textit{Battery life} & $8-10$ h & $6$ h\\
		\textit{Communication} & Wi-Fi & Radio Protocol \textdegree C \\
		\textit{Receiver} & Wi-Fi Router & Awinda Station \\
		\textit{Hardware} & $17$ wired sensors, full-body lycra suit &  $17(+1)$ wireless sensors, t-shirt + straps \\
		\textit{Charging} & USB cable & Charging Station \\ \midrule
  & \multicolumn{2}{c}{\textbf{Trackers}} \\ \midrule
            \textit{Static accuracy (Roll/Pitch) } & \multicolumn{2}{c}{$0.2$ deg} \\
		\textit{Static accuracy (heading)} & \multicolumn{2}{c}{$0.5$ deg}\\
		\textit{Dynamic accuracy}& \multicolumn{2}{c}{$1$ deg RMS} \\
		\textit{Accelerometer range} & \multicolumn{2}{c}{$\pm160$ m/s\textsuperscript{2} }  \\
    \textit{Gyroscope range} & \multicolumn{2}{c}{$\pm2000$ deg/s } \\
		\bottomrule
	\end{tabular}}
\end{table}

\begin{figure}[ht]
	\centering
	\hfill
        \begin{minipage}[c]{.7\textwidth}
	\centering
	\includegraphics[width=\textwidth,trim={0cm 7cm 16cm 0cm},clip]{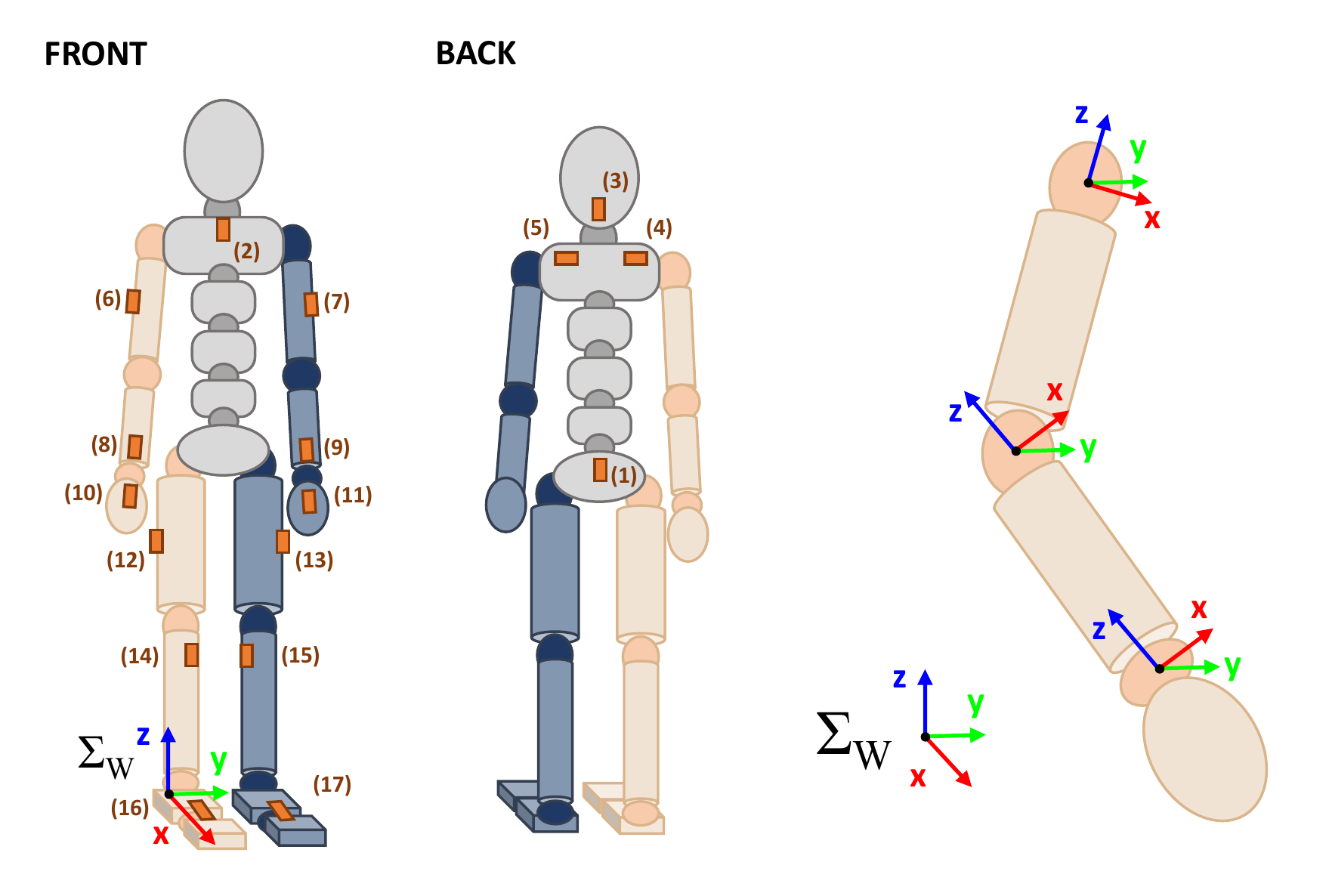}
    \caption*{(a)}
	\end{minipage}\hspace*{\fill}
 	\begin{minipage}[c]{.3\textwidth}
		\centering
		\footnotesize
			\begin{tabular}{ccc}
				\toprule
				& \emph{\textbf{Link}} & \emph{\textbf{IMUs}} \\ \midrule
				\textbf{Trunk} & \textit{Pelvis} & (1)      \\ 
				& \textit{L5}       &             \\ 
				& \textit{L3}       &             \\ 
				& \textit{T12}      &             \\ 
				& \textit{T8}       & (2)        \\ 
				\\ \midrule
				\textbf{Head}  & \textit{Neck}     &             \\ 
				& \textit{Head}     & (3)       \\ 
				\\ \midrule 
				\textbf{Arm}   & \textit{Shoulder} & (4)-(5)        \\ 
				& \textit{UpperArm} & (6)-(7)        \\ 
				& \textit{ForeArm}  & (8)-(9)       \\ 
				& \textit{Hand}     & (10)-(11)        \\ 
				\\ \midrule 
				\textbf{Leg}   & \textit{UpperLeg} & (12)-(13)       \\ 
				& \textit{LowerLeg} & (14)-(15)        \\ 
				& \textit{Foot}     & (16)-(17)       \\ 
				& \textit{Toe}      &             \\ 
				\bottomrule
			\end{tabular}
   \caption*{(b)}
	\end{minipage}\hspace*{\fill}
	\linespread{1}
	\caption{ (a) Xsens model of a human subject performing N-pose. The IMUs, the global reference frame $\Sigma_{W}$, and local reference frames for the arm, as an example, are highlighted. (b) Positions of the IMUs on the body segments.}
	\linespread{1.5}	
	\label{fig:xsensmodel}
\end{figure}

The system runs in real-time with a maximum update rate of $240$ Hz for MTx and $60$ Hz for MTw, respectively. In order to estimate the variations of the body links orientation and position, gyroscope and accelerometer signals are continuously updated and integrated within a biomechanical model of the human body (see Figure \ref{fig:xsensmodel}a). The Xsens human model includes 23 segments (pelvis, L5, L3, T12, T8, neck, head, and right and left shoulder, upper arms, forearms, hands, upper legs, lower legs, feet, and toes) linked together by 22 joints with three DoFs. The 17 IMUs are attached to such body segments as indicated in the Table in Figure \ref{fig:xsensmodel}b. 
The origins of the joints are defined in the center of the functional axes with the directions of the $\textit{x}-\textit{y}-\textit{z}$ axes being related to functional movements (see Figure \ref{fig:xsensmodel}a).

When a sensor is attached to the human body, the initial transformation between the sensor and the body segment on which is located is unknown. 
To address this issue and express the body segment kinematics in a global frame $\Sigma_{W}$, a calibration procedure must be performed. 
Then, the sensors positions and orientations can be estimated by integrating the gyroscope data and double integrating the accelerometer data in time. To achieve six full degrees of freedom (DoFs) tracking of the subject, the information coming from the sensors is translated to body segments using the above-mentioned biomechanical model and the sensor pose w.r.t. the body segment, found in the calibration step. It should be pointed out that the uncertainty of the joint position and rotation increases throughout time, due to the sensor noise related to skin and soft tissue artifacts. However, the resulting error is compensated by using the joint measurement updates, exploiting the knowledge that two segments are on average connected but with statistical uncertainty. In addition, a Kalman filter is used to correct the kinematics for both the drift and the uncertainty of the joint position. Detailed information can be found in \cite{xsensmanual}.

\noindent As mentioned in the introduction, the Xsens MVN system enables the online streaming of 3D kinematic data. Specifically, the network streamer, which supports both UDP and TCP, can provide the following information:
\begin{itemize}
    \item 3D position, velocity, acceleration, orientation (both quaternion and Euler format), angular velocity, and angular acceleration of all the twenty-three body segments w.r.t. a global (earth-fixed) coordinate system $\Sigma_{W}$, which is located at the right heel of the subject (illustrated in Figure \ref{fig:xsensmodel}).
    \item Joint angles as Euler angles ZXY (flexion/extension, abduction/adduction, internal/external rotation) for all the twenty-two joints.
    \item 3D positions of a set of virtual optical markers i.e. some points of interest corresponding to anatomical landmarks, in the links frames. 
    \item 3D position of the center of mass (CoM), based on the segment poses together with a body mass distribution model. 
    \item Character meta data, scaling data, time code and others.
\end{itemize}
 
\subsection*{Robot Operating System (ROS)}
Robot Operating System (ROS)~\cite{quigley2009ros} is an open-source framework that facilitates robot software development. Despite what its name indicates, ROS is not a traditional Operating System (OS); instead, it could be considered a middleware or a meta-operating system for robots that provides a communication structure layer above the host OS. It provides services one could expect from an OS, including hardware abstraction, low-level device control and drivers, commonly-used functionalities, message-passing communication, and package management. In addition, it comprises a collection of tools for building, writing, and running software across multiple computers and robots, enabling the development of robotic applications. Although ROS is not a real-time framework, it is possible to integrate ROS with real-time code.

The ROS runtime network (graph) is a peer-to-peer network of processes (nodes) that follows a distributed architecture, allowing for integrating multiple nodes running on different machines. In ROS, each node represents a specific functionality, such as data processing, control algorithm, or visualization. Nodes communicate with each other by passing messages over a publish-subscribe messaging system. In this sense, ROS nodes communicate with each other using a lightweight, language-independent message-passing system. Nodes can publish messages to specific topics, to which other nodes can subscribe to receive the messages. Overall, ROS implements several communication styles, including synchronous Remote-Procedure-Call (RPC)-style communication over services, asynchronous data streaming over topics, and data storage on a Parameter Server. 

In addition to the communication structure, ROS also provides code and resources organized in packages. E.g., the software package presented in this manuscript. In ROS, each package comprises a software unit of functionality. Some packages may require other packages to work (dependencies). ROS includes tools for dealing with dependencies and building and distributing packages. Moreover, ROS furnishes a hardware abstraction layer, allowing developers to write software not strictly related to specific hardware components, making reusing code and switching platforms more straightforward. It also includes visualization and debugging tools, allowing one to picture sensor data, robot states, and the interconnection of the nodes inside the graph.

ROS has gained momentum and popularity in the robotics community thanks to its flexibility, scalability, and extensive sets of software modules and packages. It is widely used in various domains, including research, industrial automation, and autonomous vehicles, among others. Thanks to the Open-Soruce philosophy behind ROS, it has a highly active community of developers and researchers that contributes to developing and maintaining ROS, shares resources, and supports other community members. As an instance of this way of thinking, we open-source this package that allows the communication between Xsens products with any other package, module, or node integrated into a ROS-based network.

\subsection*{Bridge}
As illustrated in Figure \ref{fig:framework}, an Xsens MVN system (both Link and Awinda) can stream online multiple 3D kinematic data through UDP/TCP. Such data can be read by a to-ROS-bridge module that 
\begin{inparaenum}[(i)]
    \item building on a copyright-free publicy available software to read and parse Xsens datagrams, map them into standard or custom ROS messages,
    \item generate a URDF model based on their content.
\end{inparaenum} 
The specific mapping between Xsens data and the corresponding ROS messages is illustrated in Table \ref{tab:xsensros}. 

\begin{table}[ht]
\caption{Mapping between Xsens data and ROS messages.}
\label{tab:xsensros}
\resizebox{\textwidth}{!}{\begin{tabular}{clllc}
\toprule
\multirow{2}{*}{\textbf{Xsens MVN data}} & \multicolumn{4}{c}{\textbf{ROS messages}}  \\ \cline{2-5} 
 & \multicolumn{1}{c}{Type} & \multicolumn{1}{c}{Sub-messages} & \multicolumn{1}{c}{Fields} & Dimension \\ \midrule
\begin{tabular}[c]{@{}c@{}}Segment \\ position + orientation\end{tabular}         & \multirow{3}{*}{xsens\_mvn\_ros::LinkState} & geometry\_msgs::Pose             & \begin{tabular}[c]{@{}l@{}}geometry\_msgs::Point position \\ geometry\_msgs::Quaternion orientation\end{tabular}     & 23 segments           \\
\begin{tabular}[c]{@{}c@{}}Segment velocity\\ (linear + angular)\end{tabular}     &                                             & geometry\_msgs::Twist            & \begin{tabular}[c]{@{}l@{}}geometry\_msgs::Vector3 linear\\ geometry\_msgs::Vector3 angular\end{tabular}             & 23 segments           \\
\begin{tabular}[c]{@{}c@{}}Segment acceleration\\ (linear + angular)\end{tabular} &                                             & geometry\_msgs::Accel            & \begin{tabular}[c]{@{}l@{}}geometry\_msgs::Vector3 linear\\ geometry\_msgs::Vector3 angular\end{tabular}             & 23 segments           \\
Joint angles                                                                      & sensor\_msgs::JointState                    &                                  & \begin{tabular}[c]{@{}l@{}}string{[}{]} name\\ float64{[}{]} position\end{tabular}                                   & 66 (22joints $\times$ 3DoFs) \\
\begin{tabular}[c]{@{}c@{}}Segment \\ position + orientation\end{tabular}         & geometry\_msgs::TransformStamped            & geometry\_msgs:transform         & \begin{tabular}[c]{@{}l@{}}geometry\_msgs::Vector3 translation\\ geometry\_msgs::Quaternion orientation\end{tabular} & 23 segments \\ \bottomrule         
\end{tabular}}
\end{table}
The URDF model generated from the content of the Xsens data is represented in Figure \ref{fig:xsensresults}a and includes 23 segments and 66 (22 $\times$ 3DoF) revolute joints. The graphic interface proposed in \cite{lorenzini2019real} is employed to display the URDF of the human model and corresponding reference frames through the ROS 3D visualizer RViz. Besides the human current body configuration, other information can be displayed by the interface as, for instance, the physical load on the human joints (represented by spheres super-imposed on the model and color-coded to denote a high, medium, or low level, i.e. red, orange and green sphere, respectively), the whole-body center of pressure (blue dot) or possible tools the human is working with (green driller).
The code to implement the to-ROS-bridge is a ROS package called \textit{xsens\_mvn\_ros} and is available on GitHub at \url{https://github.com/hrii-iit/xsens_mvn_ros}. The main documentation can be found at \url{https://hrii-iit.github.io/xsens_mvn_ros/index.html}.

\section*{\textsf{Results}}
\label{Sec::results}
\begin{figure}[h]
    \centering
    \includegraphics[width=\textwidth,trim={0 1cm 0 2cm},clip]{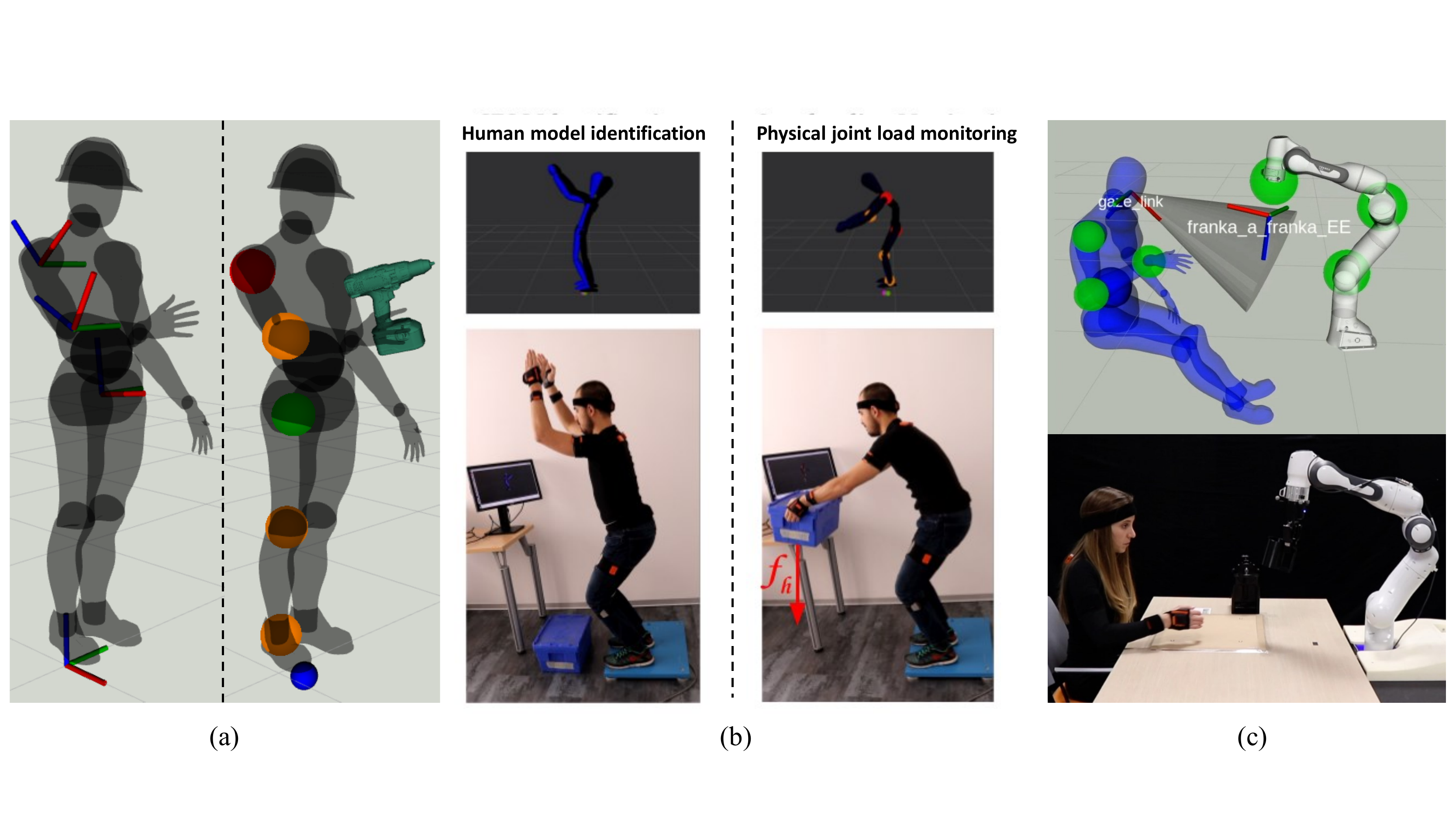}
    \caption{(a) Representation in RViz of the URDF model of the human generated from the content of the Xsens data. (b) Personalisable human ergonomics monitoring framework developed in \cite{fortini2020framework}. (c) Learning framework to maximise human-robot coefficiency proposed in \cite{lagomarsino2023maximising}. }
    \label{fig:xsensresults}
\end{figure}

As illustrated in Figure \ref{fig:framework}, once the information from Xsens MVN has been translated into ROS messages, it is possible to employ them both in simulation and in the real world. Figure \ref{fig:xsensresults} depicts a couple of Xsens-to-ROS bridge applications. A customizable human ergonomics framework was proposed in \cite{fortini2020framework} that integrates a method for online identification of a human model and an ergonomics monitoring function. In Figure \ref{fig:xsensresults}b the experiment conducted to identify the model parameters through a set of static poses and then estimate the joints overloading given by an external load is represented. Instead, in Figure \ref{fig:xsensresults}c the experiment carried out in \cite{lagomarsino2023maximising} is shown. The novel concept of human-robot coefficiency is here introduced and modelled by identifying implicit indicators of human comfort and discomfort as well as calculating the robot energy consumption in performing a desired trajectory. Then, a reinforcement learning approach is proposed that uses the human-robot coefficiency score as a reward to adapt and learn online the combination of robot interaction parameters that maximizes such coefficiency. In both the cited studies, the Xsens-to-ROS bridge is employed to collect data on human motion and map them into ROS messages to feed the related ROS nodes. 

Beyond these works, the Xsens-to-ROS bridge has been used in many other applications: humans simulations \cite{fortini2022open} and ergonomic risk assessment \cite{santopaolo2022biomechanical,lorenzini2022online,ventura2021flexible,fortini2020real,lorenzini2020online}, performance evaluation of human tracking systems \cite{fortini2023markerless} or human-robot/feedback interfaces \cite{gandarias2022enhancing,gholami2022quantitative,kim2021directional,lorenzini2022performance,kim2018ergotac}, ergonomic human-robot collaboration \cite{merlo2023ergonomic,merlo2022dynamic,kim2021human,kim2019towards,lorenzini2019new,kim2019adaptable,lorenzini2019toward,lorenzini2018synergistic}, robot control \cite{10081043,9981948,doganayidil}, assistive robotics \cite{ruiz2022improving,du23bidirectional} and teleoperation \cite{wu2019teleoperation}.

\section*{\textsf{Future works}}
\label{Sec::discussion}
The Xsens-to-ROS bridge will be soon upgraded to ROS2 \cite{macenski2022robot}. A connection to the design, development, and simulation suite Gazebo \url{https://gazebosim.org/home} will be also established.

\printbibliography

\section*{\textsf{Acknowledgments}}
This work was supported in part by the European Union’s Horizon 2020 research and innovation program under Grant Agreement No. 871237 (SOPHIA) in part by the ERC-StG Ergo-Lean (Grant Agreement No.850932).

\end{document}